\documentclass{article}

\usepackage[nonatbib,preprint]{neurips_2023_modified}
\usepackage[numbers,sort&compress]{natbib}

\usepackage[utf8]{inputenc} 
\usepackage[T1]{fontenc}    
\usepackage{url}            
\usepackage{booktabs}       
\usepackage{amsfonts}       
\usepackage{nicefrac}       
\usepackage{microtype}      

\usepackage{caption}
\usepackage{subcaption}
\usepackage{wrapfig}
\usepackage{etoolbox}
\usepackage{verbatim}
\usepackage[inline]{enumitem}
\usepackage{amsmath}
\usepackage{graphicx}
\usepackage{tabularray} 
\usepackage{caption}
\usepackage{multirow}
\usepackage{makecell}
\usepackage{bbm}

\usepackage{colortbl}

\usepackage[usenames,dvipsnames]{xcolor}
\usepackage{hyperref}
\hypersetup{ %
	pdfauthor={},
	pdftitle={},
	pdfsubject={},
        pdflang={en},
	bookmarks,
	bookmarksnumbered,
	backref=page,
	pageanchor=true,
	plainpages=false,
	colorlinks=true,%
	linktocpage=true,
	citecolor=MidnightBlue,%
	filecolor=BrickRed,%
	linkcolor=NavyBlue,%
	urlcolor=NavyBlue 
}

\usepackage{tikz}
\usetikzlibrary{arrows,positioning,shapes}
\tikzset{box/.style={rectangle, draw=black, minimum size=0.25cm}}

\usepackage{xspace}
\usepackage[title]{appendix}

\usepackage{environ}
\usepackage{esvect} 
\usepackage[ruled,vlined]{algorithm2e}
\usepackage{algcompatible}
\usepackage[noend]{algpseudocode}

\NewEnviron{boxcode2col}[4]{
    \begin{center}
        \scalebox{#3}{
            \setlength\tabcolsep{4pt}
            \renewcommand{\arraystretch}{#4}
            \begin{tabular}{|p{#1}p{#2}|}   
                \hline
                \BODY
                [1ex]
                \hline
            \end{tabular}
        }
    \end{center}
}

\NewEnviron{boxcode}[3]{
    \begin{center}
        \scalebox{#2}{
            \setlength\tabcolsep{4pt}
            \renewcommand{\arraystretch}{#3}
            \begin{tabular}{|p{#1}|}
                \hline
                \vspace{0.1mm}    
                \BODY
                \\\hline
    		\end{tabular}
    	}
    \end{center}
}

\newcommand{\textcode}[1]{{\fontfamily{cmtt}\selectfont #1}\xspace}

\newcommand{\TaskAvatar}{\textcode{\textsc{avatar}}}
\newcommand{\TaskGoal}{\textcode{\textsc{goal}}}
\newcommand{\TaskMarker}{\textcode{\textsc{marker}}}
\newcommand{\TaskFree}{\textcode{\textsc{free}}}
\newcommand{\TaskWall}{\textcode{\textsc{wall}}}

\newcommand{\TEast}{\ensuremath{\rhd}}

\newcommand{\TDSLMove}{\textcolor{blue}{\ \textcode{M}}\ }
\newcommand{\TDSLTurnLeft}{\textcolor{blue}{\ \textcode{L}}\ }
\newcommand{\TDSLTurnRight}{\textcolor{blue}{\ \textcode{R}}\ }

\newcommand{\DSLMove}{\textcode{move}}
\newcommand{\DSLTurnLeft}{\textcode{turnLeft}}

\newcommand{\DSLTurnRight}{\textcode{turnRight}}

\newcommand{\DSLPickMarker}{\textcode{pickMarker}}

\newcommand{\DSLRepeat}{\textcode{\textsc{Repeat}}}
\newcommand{\DSLRepeatUntil}{\textcode{\textsc{RepeatUntil}}}
\newcommand{\DSLIf}{\textcode{\textsc{If}}}
\newcommand{\DSLIfElse}{\textcode{\textsc{IfElse}}}

\newcommand{\DSLWhile}{\textcode{\textsc{While}}}
\newcommand{\DSLRun}{\textcode{\textsc{Run}}}
\newcommand{\DSLBoolGoal}{\textcode{goal}}
\newcommand{\DSLBoolPathAhead}{\textcode{pathAhead}}

\newcommand{\DSLBoolNoPathAhead}{\textcode{no-pathAhead}}

\newcommand{\DSLBoolMarkerPresent}{\textcode{markerPresent}}

\newcommand{\hocTypeBold}{\textbf{HoCMaze}}
\newcommand{\hocType}{\textnormal{HoCMaze}}

\newcommand{\karelTypeBold}{\textbf{Karel}}
\newcommand{\karelType}{\textnormal{Karel}}

\definecolor{TutorColour}{RGB}{105, 105, 105}
\definecolor{PCFGColour}{RGB}{204, 153, 24}

\definecolor{URLColor}{RGB}{0,103,149}

\newcommand{\TechChatGPT}{\textcode{ChatGPT}}
\newcommand{\TechGPTFour}{\textcode{GPT-4}}

\newcommand{\Evaluators}{\textcode{evals}}

\newcommand{\TaskZero}{\textcode{{T0}}}
\newcommand{\TaskOne}{\textcode{{T1}}}
\newcommand{\TaskTwo}{\textcode{{T2}}}
\newcommand{\TaskThree}{\textcode{{T3}}}
\newcommand{\TaskFour}{\textcode{{T4}}}
\newcommand{\TaskFive}{\textcode{{T5}}}
\newcommand{\TaskSix}{\textcode{{T6}}}
\newcommand{\TaskSeven}{\textcode{{T7}}}
\newcommand{\TaskEight}{\textcode{{T8}}}
\newcommand{\TaskNine}{\textcode{{T9}}}

\definecolor{ExpHighlight}{rgb}{0.8, 0.1, 0.1}
\definecolor{mygreen}{rgb}{0,0.6,0}
\definecolor{mygray}{rgb}{0.5,0.5,0.5}
\definecolor{mymauve}{rgb}{0.58,0,0.82}
\definecolor{CodeHighlight}{rgb}{1,1,0.7}
\definecolor{CodeGray}{rgb}{0.8,0.8,0.8}

\definecolor{promptinputcolor}{rgb}{0.58,0,0.82}
\newcommand{\promptheader}[1]{{\large{\textcolor{RoyalPurple}{\textbf{#1}}}}}
\newcommand{\promptinput}[1]{{\textcolor{promptinputcolor}{\textcode{#1}}}}

\title{Evaluating ChatGPT and GPT-4 \\for Visual Programming\thanks{This article is a full version of the poster (extended abstract) from ICER'23~\cite{icer23poster_ChatGPT_visualprog}.}}

\author{
    Adish Singla\\
    MPI-SWS\\
    \texttt{adishs@mpi-sws.org} 
}

\begin{document}

\maketitle

\begin{abstract}
Generative AI and large language models have the potential to drastically improve the landscape of computing education by automatically generating personalized feedback and content. Recent works have studied the capabilities of these models for different programming education scenarios; however, these works considered only text-based programming, in particular, Python programming. Consequently, they leave open the question of how well these models would perform in visual programming domains popularly used for K-8 programming education. The main research question we study is: \emph{Do state-of-the-art generative models show advanced capabilities in visual programming on par with their capabilities in text-based Python programming?} In our work, we evaluate two models, ChatGPT (based on GPT-3.5) and GPT-4, in visual programming domains for various scenarios and assess performance using expert-based annotations. In particular, we base our evaluation using reference tasks from the domains of \emph{Hour of Code: Maze Challenge} by Code.org and Karel. Our results show that these models perform poorly and struggle to combine spatial, logical, and programming skills crucial for visual programming. These results also provide exciting directions for future work on developing techniques to improve the performance of generative models in visual programming.
\end{abstract}
%

\vspace{-3mm}
\section{Introduction}\label{sec.introduction}
\vspace{-1mm}
Generative AI and large language models (LLMs) hold great promise in enhancing computing education by powering next-generation educational technologies for introductory programming. In particular, this potential lies in the advanced capabilities of state-of-the-art models---like  OpenAI's ChatGPT~\cite{ChatGPT} and GPT-4~\cite{GPT4}---to automatically generate high-quality personalized feedback and content~\cite{DBLP:journals/corr/abs-2303-12712,baidoo2023education,LIM2023100790}. In our work, we seek to investigate the capabilities of these models in visual programming domains popularly used for K-8 programming education, including domains like Scratch~\cite{DBLP:journals/cacm/ResnickMMREBMRSSK09}, \emph{Hour of Code: Maze Challenge} by Code.org~\cite{hourofcode_maze,codeorg}, and Karel~\cite{pattis1995karel,karel_stanford_course_new,intro_to_karel_codehs}.

Recent works have studied the capabilities of these models for various programming education scenarios such as program repair, hint generation, pair programming, and task synthesis~\cite{icer23poster_ChatGPT_pythonprog,DBLP:conf/icer/SarsaDH022,edm23-pyfixv,leinonen23sigcse,CopilotWeb,DBLP:journals/corr/abs-2210-14306,DBLP:conf/icse/Imai22,DBLP:journals/corr/abs-2306-05153}. A study in 2022 had ranked OpenAI's Codex (based on GPT-3)~\cite{DBLP:journals/corr/abs-2107-03374} in the top quartile w.r.t students in a large Python programming course~\cite{DBLP:conf/ace/Finnie-AnsleyDB22}. A recent study in contemporary work has shown that OpenAI's GPT-4 drastically outperforms ChatGPT (based on GPT-3.5) and comes close to human tutors' performance for several scenarios~\cite{icer23poster_ChatGPT_pythonprog}. However, these above-mentioned works have considered only text-based (Python) programming and leave open the question of how well these models perform in visual programming domains. The main research question we study is: 
\vspace{-1mm}
\begin{quote}
\emph{Do state-of-the-art generative models show advanced capabilities in visual programming on par with their capabilities in text-based Python programming?}
\end{quote}
\vspace{3mm}

In our work, we evaluate two models, ChatGPT (based on GPT-3.5) and GPT-4, in visual programming domains for a variety of scenarios. These scenarios are designed to evaluate various generative and problem-solving capabilities of these models in visual programming. More concretely, we consider the following three scenarios: (i) \emph{execution trace}; (ii) \emph{solution synthesis}; (iii) \emph{task synthesis}. We provide further details about these scenarios in the following sections.

We evaluate the performance of different methods using expert-based annotations involving a mix of quantitative and qualitative assessments. We base our evaluation using ten reference tasks from the visual programming domains of \emph{Hour of Code: Maze Challenge} by Code.org~\cite{hourofcode_maze,codeorg} and  Karel~\cite{pattis1995karel,karel_stanford_course_new,intro_to_karel_codehs}. Our results show that GPT-4 drastically improves up on ChatGPT (based on GPT-3.5); however, the performance of GPT-4 is still quite poor as it struggles to combine spatial, logical, and programming skills crucial for visual programming.

The rest of this paper is organized as follows. Section~\ref{sec.problemsetup} provides an overview of our evaluation setup. Sections~\ref{sec.executiontrace},~\ref{sec.solutionsynthesis},~and~\ref{sec.tasksynthesis} provide results for the above-mentioned three scenarios. Section~\ref{sec.conclusion} discusses some limitations of our current work and directions for future work.


\section{Evaluation Setup}\label{sec.problemsetup}
This section provides an overview of our evaluation setup, including the scenarios, visual programming domains, and the overall process used for evaluation.

\paragraph{Evaluation scenarios.} 
In our work, we consider the following three scenarios that capture various generative and problem-solving capabilities of LLMs in visual programming:
\begin{enumerate}[label={(\roman*)},leftmargin=16pt,parsep=4pt]
    \item \emph{Execution trace}, i.e., analyzing the behavior when executing a given code on a task, motivated by the application of analyzing students' attempts to provide feedback~\cite{DBLP:conf/edm/Marwan0MCBP21,edm22-student-synthesis,ProgresSyn2023}.
    \item \emph{Solution synthesis}, i.e., generating solution codes for a given task, motivated by the application of completing students' partial programs or giving next-step hints~\cite{DBLP:conf/lats/PiechSHG15,DBLP:conf/aied/PriceZB17,edm20-zero-shot}.
    \item \looseness-1\emph{Task synthesis}, i.e., generating new tasks that exercise specific concepts, motivated by the application of providing new practice tasks to students in visual programming domains~\cite{DBLP:conf/nips/AhmedCEFGRS20,DBLP:conf/aied/GhoshTDS22,neurtasksyn}.
\end{enumerate}

\begin{figure*}[t!]
\centering
    \scalebox{0.9}{
    \setlength\tabcolsep{8.0pt}
    \renewcommand{\arraystretch}{1.2}
    \begin{tabular}{c|c|ccl|l}
        \toprule
        \multicolumn{1}{c|}{\textbf{Task ID}} & \multicolumn{1}{c|}{\textbf{Domain}} & \multicolumn{3}{c|}{\textbf{Complexity of Solution Code}} & \multicolumn{1}{c}{\textbf{Source}} \\
         &   & \multicolumn{1}{c}{Size} & \multicolumn{1}{c}{Depth} & \multicolumn{1}{c|}{Structure} &  \\        
        \midrule
        \TaskZero{} & \hocType{} & $6$ & $1$ & \text{{\{\DSLRun{}\}}} & {\hocType{}:4~\cite{hourofcode_maze}} \\
        \TaskOne{} & \hocType{} & $4$ & $2$ & \text{{\{\DSLRun\{\DSLRepeat{}\}\}}} & {\hocType{}:7~\cite{hourofcode_maze}} \\
        \TaskTwo{} & \hocType{} & $6$ & $2$ & \text{{\{\DSLRun\{\DSLRepeatUntil{}\}\}}} & {\hocType{}:12~\cite{hourofcode_maze}} \\
        \TaskThree{} & \hocType{} & $5$ & $3$ & \text{{\{\DSLRun\{\DSLRepeatUntil\{\DSLIf{}\}\}\}}} & {\hocType{}:16~\cite{hourofcode_maze}} \\
        \TaskFour{} & \hocType{} & $5$ & $3$ & \text{{\{\DSLRun\{\DSLRepeatUntil\{\DSLIfElse{}\}\}\}}} & {\hocType{}:18~\cite{hourofcode_maze}} \\
        \TaskFive{} & \karelType{} & $6$ & $1$ & \text{{\{\DSLRun{}\}}} & {\karelType{}:OurFirst~\cite{intro_to_karel_codehs}} \\
        \TaskSix{} & \karelType{} & $4$ & $2$ & \text{{\{\DSLRun\{\DSLRepeat{}\}\}}} & {\karelType{}:RowOfBalls} \\
        \TaskSeven{} & \karelType{} & $8$ & $2$ & \text{{\{\DSLRun\{\DSLWhile{}\}\}}} & {\karelType{}:Diagonal~\cite{intro_to_karel_codehs}} \\
        \TaskEight{} & \karelType{} & $6$ & $3$ & \text{{\{\DSLRun\{\DSLRepeat\{\DSLIf{}\}\}\}}} & {\karelType{}:Opposite~\cite{intro_to_karel_codehs}} \\
        \TaskNine{} & \karelType{} & $8$ & $3$ & \text{{\{\DSLRun\{\DSLWhile\{\DSLIf{}\}\}\}}} & {\karelType{}:Stairway~\cite{intro_to_karel_codehs}} \\
        \bottomrule
    \end{tabular}
    }
    \caption{Summary of ten reference tasks used in our evaluation. We have five tasks each from the visual programming domains of \emph{Hour of Code: Maze Challenge} by Code.org (in short, \hocType{})~\cite{hourofcode_maze,codeorg} and  Karel~\cite{pattis1995karel,karel_stanford_course_new,intro_to_karel_codehs}. Figures~\ref{fig.problemsetup.task.hoc}~and~\ref{fig.problemsetup.task.karel} provide an illustration of tasks \TaskFour{} and \TaskNine{}, respectively.}
    \label{fig.problemsetup.dataset}
\end{figure*}

\begin{figure*}[t!]
\centering
    \begin{subfigure}[b]{0.62\textwidth}
        \centering
        \includegraphics[height=3.44cm]{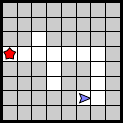}
        \caption{Task's 8x8 visual grid}
        \label{fig.solution_synthesis.example.hoc.task} 
    \end{subfigure}
    \quad    
    \begin{subfigure}[b]{0.3\textwidth}
        \centering
        \includegraphics[height=3.5cm]{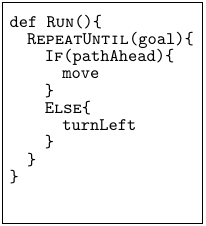}
        \caption{Minimal-sized solution code}
        \label{fig.solution_synthesis.example.hoc.code}
    \end{subfigure}
    %
    \caption{\looseness-1This figure shows \emph{HoCMaze:18} task from the \hocType{} domain referred to as \TaskFour{} in Figure~\ref{fig.problemsetup.dataset}~\cite{hourofcode_maze}. \textbf{(a)} shows the task's  8x8 visual grid and \textbf{(b)} shows the minimal-sized solution code for this task. The task's visual grid comprises the following elements: \TaskAvatar{} (purple dart), \TaskGoal{} (red star), \TaskFree{} cells (white-colored grid cells), and \TaskWall{} cells (gray-colored grid cells). When solving this task, the objective is to combine available code blocks for navigating the \TaskAvatar{} to the \TaskGoal{}. Importantly, there is also an upper limit on the number of blocks that can be used in a solution code (typically, this limit is set to be the size of the minimal solution code). The minimal-sized solution for this task uses a total of $5$ blocks, i.e., \DSLRun{}, \DSLRepeatUntil\textcode{(\DSLBoolGoal{})}, \DSLIfElse{}\textcode{(\DSLBoolPathAhead{})}, \DSLMove{}, \DSLTurnLeft{}.}
    \label{fig.problemsetup.task.hoc}
\end{figure*}

\begin{figure*}[t!]
\centering
    \begin{subfigure}[b]{0.62\textwidth}
        \centering
        \includegraphics[height=3.44cm]{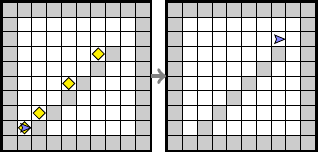}
        \caption{Task's 10x10 visual pregrid (left) and 10x10 visual postgrid (right)}
        \label{fig.solution_synthesis.example.karel.task} 
    \end{subfigure}
    \quad
    \begin{subfigure}[b]{0.3\textwidth}
        \centering
        \includegraphics[height=3.5cm]{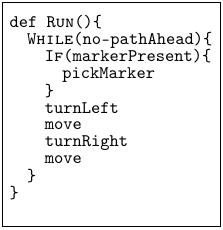}
        \caption{Minimal-sized solution code}
        \label{fig.solution_synthesis.example.karel.code}
    \end{subfigure}
    %
    \caption{This figure shows \emph{Stairway} task from the \karelType{} domain referred to as \TaskNine{} in Figure~\ref{fig.problemsetup.dataset}~\cite{intro_to_karel_codehs}. As considered in the work of \cite{neurtasksyn}, we use only a single pregrid-postgrid task specification for Karel in our evaluation -- this simplifies the task representation in prompts and keeps the overall evaluation setting for \karelType{} and \hocType{} domains similar. \textbf{(a)} shows the task's a pair of 10x10 visual pregrid and postgrid and \textbf{(b)} shows the minimal-sized solution code for this task. The task's pregrid and postgrid comprise the following elements: \TaskAvatar{} (purple dart), \TaskMarker{} objects (yellow diamonds), \TaskFree{} cells (white-colored grid cells), and \TaskWall{} cells (gray-colored grid cells). When solving this task, the objective is to combine available code blocks that transform the pregrid to postgrid. Importantly, similar to the \hocType{} domain, we also consider an upper limit on the number of blocks that can be used in a solution code (set to be the size of the minimal solution code)~\cite{neurtasksyn}. The minimal-sized solution for this task uses a total of $8$ blocks, i.e., \DSLRun{}, \DSLWhile\textcode{(\DSLBoolNoPathAhead{})}, \DSLIf{}\textcode{(\DSLBoolMarkerPresent{})}, \DSLPickMarker{}, \DSLTurnLeft{}, \DSLMove{}, \DSLTurnRight{}, \DSLMove{}.}
    \label{fig.problemsetup.task.karel}
\end{figure*}

\paragraph{Visual programming domains and tasks.} We base our evaluation using ten reference tasks from the visual programming domains of \emph{Hour of Code: Maze Challenge} by Code.org (in short, \hocType{})~\cite{hourofcode_maze,codeorg} and  Karel~\cite{pattis1995karel,karel_stanford_course_new,intro_to_karel_codehs}. Figure~\ref{fig.problemsetup.dataset} provides information about these reference tasks in terms of complexity and programming concepts exercised; Figures~\ref{fig.problemsetup.task.hoc}~and~\ref{fig.problemsetup.task.karel} show an illustrative task from \hocType{} and \karelType{} domains, respectively. These tasks are typically suitable for elementary-level programming education and variants of these tasks have been extensively used in literature~\cite{DBLP:conf/lats/PiechSHG15,edm20-zero-shot,DBLP:conf/nips/AhmedCEFGRS20,DBLP:conf/aied/GhoshTDS22,neurtasksyn}. Each of these reference tasks has a unique minimal-sized solution code and the task complexity can be captured through the properties of this solution code. As shown in Figure~\ref{fig.problemsetup.dataset}, we characterize a task and its solution code through the following properties: (a) \emph{size} is the number of code ``blocks'' in the solution code (i.e., code tokens corresponding to environment actions or programming constructs like loops/condition); (b) \emph{depth} is the depth of the Abstract Syntax Tree representation of the solution code; (c) \emph{structure} is the nesting structure of programming constructs in the solution code. We refer the reader to \cite{neurtasksyn} for a more formal specification about the space of tasks and codes in these visual programming domains.

\paragraph{Methods evaluated.} We evaluate two methods in our work: (a) \TechChatGPT{} that uses OpenAI's ChatGPT (based on GPT-3.5) as its LLM via web platform~\cite{ChatGPT,ChatGPTWeb}; (b) \TechGPTFour{} that uses OpenAI's GPT-4 as its LLM via web platform~\cite{GPT4,GPT4Web}. Prompts used to interact with LLMs are provided in the subsequent sections for different scenarios. Next, we describe the interaction process with these models and outputs for evaluation. For a given method and scenario, we have $10$ total instances for evaluation corresponding to $10$ reference tasks. First, we manually perform $n_{\text{queries}}=5$ queries to an LLM through the web platform to generate multiple outputs per instance; then, we manually select one of these outputs as the final output that performs best in terms of scenario-specific metrics. We describe further scenario-specific details in the subsequent sections.

\paragraph{Metrics and evaluation process.} We will introduce scenario-specific performance metrics in the subsequent sections. Even though the performance metrics used in our work can be objectively assessed, it is challenging to fully automate their assessment. Hence, we assess performance using expert-based annotation as typically done in the literature~\cite{DBLP:conf/aied/PriceZB17,DBLP:conf/aied/GhoshTDS22,neurtasksyn}. In particular, we have $n_{\text{\Evaluators{}}}=1$ human expert evaluator who provides annotations to assess the quality of generated output for each instance w.r.t. corresponding performance metrics. Then, for each method, we report aggregated results averaged across $10$ instances.
%


\section{Execution Trace Scenario}\label{sec.executiontrace}
This section is dedicated to the scenario of \emph{execution trace}, i.e., analyzing the behavior when executing a given code on a task. This scenario is motivated by the application of analyzing students' attempts to provide feedback~\cite{DBLP:conf/edm/Marwan0MCBP21,edm22-student-synthesis,ProgresSyn2023}. Next, we provide details of this scenario's prompt, input-output formats, performance metrics, and results.

\begin{figure}[t!]
    \centering
    \scalebox{0.90}{
        \setlength\tabcolsep{5pt}
        \renewcommand{\arraystretch}{1.2}
\begin{tabular}{||p{0.99\linewidth}||}
    \hline
    \multicolumn{1}{||c||}{\promptheader{Prompt: Execution Trace (\hocTypeBold{})}} \\ 
    I am learning to code using the block-based visual programming domain of Hour of Code: Maze Challenge from code.org. 
    \newline
    \newline
    In this domain, the following types of coding blocks are available. 
    \newline
    - \quad Basic action blocks: move forward, turn left, turn right.
    \newline
    - \quad Boolean conditions: path ahead, path to the left, path to the right.
    \newline
    - \quad Loops: repeatUntil(goal)\{\}, repeat(int)\{\}.
    \newline
    - \quad Conditionals: If(boolean)\{\}, If(boolean)\{\}Else\{\}.
    \newline
    \newline
    In this domain, a task is represented as an 8x8 visual grid that contains WALL cells, FREE cells, AVATAR (with specific location and direction), and GOAL. We represent a task's 8x8 visual grid with the following symbols.
    \newline  
    - \quad \# represents a WALL cell.
    \newline
    - \quad + represents a FREE cell.
    \newline    
    - \quad * represents GOAL. 
    \newline    
    - \quad E represents AVATAR's location facing East direction.
    \newline    
    - \quad W represents AVATAR's location facing West direction.
    \newline    
    - \quad N represents AVATAR's location facing North direction.
    \newline    
    - \quad S represents AVATAR's location facing South direction.
    \newline
    \newline  
    Below I am giving you a task and its solution code. A solution code for a task takes AVATAR to GOAL when executed.
    \newline
    \newline    
    --- Task ---
    \newline    
    \promptinput{\{grid\_ascii\_representation\}}    
    \newline
    \newline
    --- Solution ---
    \newline    
    \promptinput{\{solution\_code\}}
    \newline
    \newline
    Can you produce an execution trace of this code on the task and tell me the sequence of AVATAR's positions, i.e., location and direction? Recall that a solution code for a task takes AVATAR to GOAL when executed. In this task, AVATAR's initial position is ((row=\promptinput{\{avatar\_row\}}, col=\promptinput{\{avatar\_col\}}), \promptinput{\{avatar\_dir\}}), and GOAL is at (row=\promptinput{\{goal\_row\}}, col=\promptinput{\{goal\_col\}}). Note that AVATAR can only move on FREE cells and will crash if it tries to go to a WALL cell.     
    \\
    \hline
\end{tabular}

    }
    \caption{Prompt for the execution trace scenario in the \hocType{} domain. This prompt has several \promptinput{placeholders} to include details for the input task and solution code. Details are in Section~\ref{sec.executiontrace}.
    }
    \label{fig.execution_trace.prompt_hoc}
\end{figure}

\begin{figure*}[t!]
    \centering
    \scalebox{1.0}{
        \setlength\tabcolsep{4.5pt}
        \renewcommand{\arraystretch}{1.4}
\begin{tabular}{l||c|cccc}
    \toprule
    \multicolumn{1}{c||}{\textbf{Method}} & \multicolumn{5}{c}{\textbf{Metrics}}\\
     & \multicolumn{1}{c|}{Overall} & \multicolumn{1}{c}{TraceCorrect} & \multicolumn{1}{c}{TracePartialCorrect} & \multicolumn{1}{c}{TransitionsCorrect} & \multicolumn{1}{c}{SensingCorrect} \\  
    \midrule
    \TechChatGPT & $10.0$ & $10.0$ & $30.0$ & $10.0$ & $10.0$ \\
    \TechGPTFour & $60.0$ & $60.0$ & $70.0$ & $90.0$ & $60.0$ \\
    \bottomrule
\end{tabular}


    }
    \caption{Results for the execution trace scenario. Details are in Section~\ref{sec.executiontrace}.}
    \label{fig.execution_trace.results}
\end{figure*}

\begin{figure*}[t!]
\centering
    \begin{subfigure}[b]{0.49\textwidth}
        \centering
            \begin{minipage}{0.49\textwidth}
	       \includegraphics[height=3.44cm]{./figs/2_problemsetup/files/fig_originaltask_hoc18.pdf}
		\end{minipage} 
		\begin{minipage}{0.49\textwidth}	
			\centering
            \includegraphics[height=3.50cm]{./figs/2_problemsetup/files/fig_originalcode_hoc18.pdf}
		\end{minipage}
        \caption{Input: Task and solution code}
        \label{fig.execution_trace.example.hoc.input}
    \end{subfigure}
    \ \ 
    %
    \begin{subfigure}[b]{.49\textwidth}
        \centering
        \includegraphics[height=3.50cm]{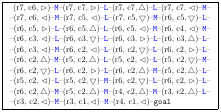}
        \caption{Output by \TechGPTFour{}}
        \label{fig.execution_trace.example.hoc.output}
    \end{subfigure}
    %
    \caption{Illustrative example for the execution trace scenario based on \emph{HoCMaze:18} task in Figure~\ref{fig.problemsetup.task.hoc}. This example highlights the struggles by \TechGPTFour{} in generating an execution trace. \textbf{(a)} shows the task and solution code provided as input. \textbf{(b)} shows the execution trace generated by \TechGPTFour{} as output. The trace is represented as a sequence of \TaskAvatar{}'s positions $(row, col, dir)$ and actions executed from the set $\{\text{\DSLMove{}} \text{\TDSLMove{}}, \text{\DSLTurnLeft{}} \text{\TDSLTurnLeft{}}, \text{\DSLTurnRight{}} \text{\TDSLTurnRight{}}\!\!\}$. Trace starts correctly with \TaskAvatar{}'s initial position of (r7, c6, \TEast{}), i.e., row 7, column 6, facing East. Moreover, the trace seemingly ends with  \TaskAvatar{} reaching the \TaskGoal{}. However, upon closer examination, we can see that the trace is incoherent, and \TaskAvatar{} crashes into \TaskWall{} cells.}
    \label{fig.execution_trace.example.hoc}
\end{figure*}

\paragraph{Prompt and output generation.} We begin by describing the content provided as input to a method and the desired output content we seek to generate. For this scenario, the input consists of a \emph{task} and \emph{solution code}; the desired output consists of an \emph{execution trace} as a sequence of \TaskAvatar{}'s positions when executing the code on the task. Figure~\ref{fig.execution_trace.prompt_hoc} shows the prompt---with placeholders for the inputs---used to interact with LLMs for the \hocType{} domain; Figure~\ref{fig.execution_trace.prompt_karel} in the appendix shows the prompt for the \karelType{} domain. The prompt starts with an overview about the domain, followed by inputs, and then summarizes the desired output. When interacting with LLMs, we first generate content using this prompt and then manually extract the execution trace as the final output for evaluation.

\paragraph{Output quality and performance metrics.} We assess the generated output along several quality attributes and use aggregated results over these quality attributes as performance metrics in our evaluation. All attributes for this scenario are binary, with a value of $1$ being better. \emph{TraceCorrect}  captures whether the generated execution trace is fully correct, i.e., it matches the trace obtained by executing the code on the task. \emph{TracePartialCorrect} relaxes the correctness criterion and captures whether the generated execution trace is partially correct, i.e., a few modifications to the generated trace would make it fully correct. \emph{TransitionsCorrect} captures whether the transitions in \TaskAvatar{}'s positions are always correct, i.e., a new position in the sequence matches what would be obtained by applying the executed action to the current position. \emph{SensingCorrect} captures whether the values of Boolean conditions (e.g., \DSLBoolGoal{} and \DSLBoolPathAhead{} for the code in Figure~\ref{fig.execution_trace.example.hoc.input}) are always correct, i.e., the action executed at any time matches what would be obtained by following the code branch based on the correct value of the Boolean condition. \emph{Overall} is $1$ when the three quality attributes of \emph{TraceCorrect}, \emph{TransitionsCorrect}, and \emph{SensingCorrect} are $1$. Human evaluators manually annotate the quality of generated output for each of the $10$ instances as mentioned in Section~\ref{sec.problemsetup}.

\paragraph{Results.} Figure~\ref{fig.execution_trace.results} provide results for various metrics aggregated across $10$ instances, reported in terms of \%. Next, we summarize some of the key findings. First, results in Figure~\ref{fig.execution_trace.results} for the metric \emph{Overall} highlight that \TechGPTFour{} ($60.0$) has substantially improved w.r.t. \TechChatGPT{} ($10.0$); in particular, this is because of improvements in spatial transitions and sensing as captured by metrics \emph{TransitionsCorrect} and \emph{SensingCorrect}. Second, these results also highlight that \TechGPTFour{} still struggles in visual programming as it achieves only $60.0$ overall performance for these elementary-level tasks. Figure~\ref{fig.execution_trace.example.hoc} provide an illustrative example highlighting the struggles by \TechGPTFour{} in generating an execution trace for \emph{HoCMaze:18} task.


\section{Solution Synthesis Scenario}\label{sec.solutionsynthesis}
This section is dedicated to the scenario of \emph{solution synthesis}, i.e., generating solution codes for a given task. This scenario is motivated by the application of completing students' partial programs or giving next-step hints~\cite{DBLP:conf/lats/PiechSHG15,DBLP:conf/aied/PriceZB17,edm20-zero-shot}. Next, we provide details of this scenario's prompt, input-output formats, performance metrics, and results.

\paragraph{Prompt and output generation.} We begin by describing the content provided as input to a method and the desired output content we seek to generate. For this scenario, the input consists of a \emph{task}; the desired output consists of a minimal-sized \emph{solution code} for the task. Figure~\ref{fig.solution_synthesis.prompt_hoc} shows the prompt---with placeholders for the inputs---used to interact with LLMs for the \hocType{} domain; Figure~\ref{fig.solution_synthesis.prompt_karel} in the appendix shows the prompt for the \karelType{} domain. The prompt starts with an overview about the domain, followed by inputs, and then summarizes the desired output. When interacting with LLMs, we first generate content using this prompt and then manually extract the solution code as the final output for evaluation.

\begin{figure}[t!]
    \centering
    \scalebox{0.90}{
        \setlength\tabcolsep{5pt}
        \renewcommand{\arraystretch}{1.2}
\begin{tabular}{||p{0.99\linewidth}||}
    \hline
    \multicolumn{1}{||c||}{\promptheader{Prompt: Solution Synthesis (\hocTypeBold{})}} \\ 
    I am learning to code using the block-based visual programming domain of Hour of Code: Maze Challenge from code.org. 
    \newline
    \newline
    In this domain, the following types of coding blocks are available. 
    \newline
    - \quad Basic action blocks: move forward, turn left, turn right.
    \newline
    - \quad Boolean conditions: path ahead, path to the left, path to the right.
    \newline
    - \quad Loops: repeatUntil(goal)\{\}, repeat(int)\{\}.
    \newline
    - \quad Conditionals: If(boolean)\{\}, If(boolean)\{\}Else\{\}.
    \newline
    \newline
    In this domain, a task is represented as an 8x8 visual grid that contains WALL cells, FREE cells, AVATAR (with specific location and direction), and GOAL. We represent a task's 8x8 visual grid with the following symbols.
    \newline  
    - \quad \# represents a WALL cell.
    \newline
    - \quad + represents a FREE cell.
    \newline    
    - \quad * represents GOAL. 
    \newline    
    - \quad E represents AVATAR's location facing East direction.
    \newline    
    - \quad W represents AVATAR's location facing West direction.
    \newline    
    - \quad N represents AVATAR's location facing North direction.
    \newline    
    - \quad S represents AVATAR's location facing South direction.
    \newline
    \newline  
    Below I am giving you a task.
    \newline
    \newline    
    --- Task ---
    \newline    
    \promptinput{\{grid\_ascii\_representation\}}    
    \newline
    \newline
    Can you generate a solution code for this task that uses the minimum number of blocks? A solution code for a task takes AVATAR to GOAL when executed. In this task, AVATAR's initial position is ((row=\promptinput{\{avatar\_row\}}, col=\promptinput{\{avatar\_col\}}), \promptinput{\{avatar\_dir\}}), and GOAL is at (row=\promptinput{\{goal\_row\}}, col=\promptinput{\{goal\_col\}}). Note that AVATAR can only move on FREE cells and will crash if it tries to go to a WALL cell. 
    \newline
    \newline   
    --- Solution ---
    \newline    
    \\
    \hline
\end{tabular}

    }
    \caption{Prompt for the solution synthesis scenario in the \hocType{} domain. This prompt has several \promptinput{placeholders} to include details for the input task. Details are in Section~\ref{sec.solutionsynthesis}.
    }
    \label{fig.solution_synthesis.prompt_hoc}
\end{figure}

\begin{figure*}[t!]
    \centering
    \scalebox{1.0}{
        \setlength\tabcolsep{4.5pt}
        \renewcommand{\arraystretch}{1.4}        
\begin{tabular}{l||c|ccccc}
    \toprule
    \multicolumn{1}{c||}{\textbf{Method}} & \multicolumn{6}{c}{\textbf{Metrics}}\\
     & \multicolumn{1}{c|}{Overall} & \multicolumn{1}{c}{SyntaxCorrect} & \multicolumn{1}{c}{CodeSolvesTask} & \multicolumn{1}{c}{CodeSimilar} & \multicolumn{1}{c}{CodeSize} & \multicolumn{1}{c}{CodeDepth} \\  
    \midrule
    \TechChatGPT & $10.0$ & $100.0$ & $50.0$ & $10.0$ & $10.0$ & $20.0$ \\
    \TechGPTFour & $40.0$ & $100.0$ & $30.0$ & $70.0$ & $40.0$ & $70.0$ \\
    \bottomrule
\end{tabular}
    }
    \caption{Results for the solution synthesis scenario. Details are in Section~\ref{sec.solutionsynthesis}.}
    \label{fig.solution_synthesis.results}
\end{figure*}

\begin{figure*}[t!]
\centering
    \begin{subfigure}[b]{0.49\textwidth}
        \centering
        \includegraphics[height=3.44cm]{./figs/2_problemsetup/files/fig_originaltask_hoc18.pdf}
        \caption{Input: Task}
        \label{fig.solution_synthesis.example.hoc.input} 
    \end{subfigure}
    \ \    
    \begin{subfigure}[b]{0.49\textwidth}
        \centering
        \includegraphics[height=3.50cm]{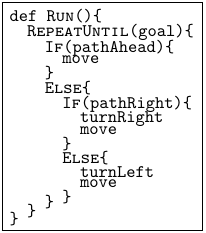}
        \caption{Output by \TechGPTFour{}}
        \label{fig.solution_synthesis.example.hoc.output}
    \end{subfigure}
    %
    \caption{\looseness-1Illustrative example for the solution synthesis scenario based on \emph{HoCMaze:18} task in Figure~\ref{fig.problemsetup.task.hoc}. This example highlights the struggles by \TechGPTFour{} in generating a solution code. \textbf{(a)} shows the task provided as input. \textbf{(b)} shows the solution code generated by \TechGPTFour{} as output. The generated code solves the input task; however, it is unnecessarily complex. In particular, it uses $9$ blocks and has depth of $4$; in contrast, the minimal-sized solution code in Figure~\ref{fig.solution_synthesis.example.hoc.code} uses only $5$ blocks and has depth of $3$.}
    \label{fig.solution_synthesis.example.hoc}
\end{figure*}

%
\paragraph{Output quality and performance metrics.} We assess the generated output along several quality attributes and use aggregated results over these quality attributes as performance metrics in our evaluation. All attributes for this scenario are binary, with a value of $1$ being better. \emph{SyntaxCorrect} captures whether the syntax of the generted code is correct in terms of programming structure and coding blocks used w.r.t. the underlying Domain Specific Language (DSL)~\cite{neurtasksyn}. \emph{CodeSolvesTask} captures whether the generated code correctly solves the input task. \emph{CodeSimilar} captures whether the generated code is similar to the minimized-sized solution code for the task, i.e., a few edits to the generated code recovers the minimized-size solution code. \emph{CodeSize} and \emph{CodeDepth} attributes capture whether the generated code has size and depth at most that of the minimal-sized solution code, respectively. \emph{Overall} is $1$ when the following holds: (i) \emph{SyntaxCorrect} attribute is $1$; (ii) at least one of the \emph{CodeSolvesTask} or \emph{CodeSimilar} attributes are $1$; (iii) \emph{CodeSize} attribute is $1$; (iv) \emph{CodeDepth} attribute is $1$. Human evaluators manually annotate the quality of generated output for each of the $10$ instances as mentioned in Section~\ref{sec.problemsetup}.

\paragraph{Results.} Figure~\ref{fig.solution_synthesis.results} provide results for various metrics aggregated across $10$ instances, reported in terms of \%. Next, we summarize some of the key findings. First, results in Figure~\ref{fig.solution_synthesis.results} for the metric \emph{Overall} highlight that \TechGPTFour{} ($40.0$) has improved w.r.t. \TechChatGPT{} ($10.0$); in particular, this is because the codes generated by \TechGPTFour{} are more similar to minimal-sized solutions as captured by metrics \emph{CodeSimilar}, \emph{CodeSize}, and \emph{CodeDepth}. In fact, even though \TechChatGPT{} performs better on the metric \emph{CodeSolvesTask}, it tends to produce a generic, complex code that can solve many tasks in the \hocType{} domain but ignores the specific task provided as input. Second, these results also highlight that \TechGPTFour{} still struggles in solving elementary-level visual programming tasks as it achieves only $40.0$ overall performance. Figure~\ref{fig.solution_synthesis.example.hoc} provide an illustrative example highlighting the struggles by \TechGPTFour{} in generating a minimal-sized solution code for \emph{HoCMaze:18} task.

%

\section{Task Synthesis Scenario}\label{sec.tasksynthesis}
This section is dedicated to the scenario of \emph{task synthesis}, i.e., generating new tasks that exercise specific concepts. This scenario is  motivated by the application of providing new practice tasks to students in visual programming domains~\cite{DBLP:conf/nips/AhmedCEFGRS20,DBLP:conf/aied/GhoshTDS22,neurtasksyn}. Next, we provide details of this scenario's prompt, input-output formats, performance metrics, and results.

\paragraph{Prompt and output generation.} We begin by describing the content provided as input to a method and the desired output content we seek to generate. For this scenario, the input consists of a \emph{solution code}; the desired output consists of a \emph{task} that would be solved by the code. Figure~\ref{fig.task_synthesis.prompt_hoc} shows the prompt---with placeholders for the inputs---used to interact with LLMs for the \hocType{} domain; Figure~\ref{fig.task_synthesis.prompt_karel} in the appendix shows the prompt for the \karelType{} domain. The prompt starts with an overview about the domain, followed by inputs, and then summarizes the desired output. When interacting with LLMs, we first generate content using this prompt and then manually extract the task as the final output for evaluation.

\begin{figure}[t!]
    \centering
    \scalebox{0.90}{
        \setlength\tabcolsep{5pt}
        \renewcommand{\arraystretch}{1.2}
\begin{tabular}{||p{0.99\linewidth}||}
    \hline
    \multicolumn{1}{||c||}{\promptheader{Prompt: Task Synthesis (\hocTypeBold{})}} \\ 
    I am learning to code using the block-based visual programming domain of Hour of Code: Maze Challenge from code.org. 
    \newline
    \newline
    In this domain, the following types of coding blocks are available. 
    \newline
    - \quad Basic action blocks: move forward, turn left, turn right.
    \newline
    - \quad Boolean conditions: path ahead, path to the left, path to the right.
    \newline
    - \quad Loops: repeatUntil(goal)\{\}, repeat(int)\{\}.
    \newline
    - \quad Conditionals: If(boolean)\{\}, If(boolean)\{\}Else\{\}.
    \newline
    \newline
    In this domain, a task is represented as an 8x8 visual grid that contains WALL cells, FREE cells, AVATAR (with specific location and direction), and GOAL. We represent a task's 8x8 visual grid with the following symbols.
    \newline  
    - \quad \# represents a WALL cell.
    \newline
    - \quad + represents a FREE cell.
    \newline    
    - \quad * represents GOAL. 
    \newline    
    - \quad E represents AVATAR's location facing East direction.
    \newline    
    - \quad W represents AVATAR's location facing West direction.
    \newline    
    - \quad N represents AVATAR's location facing North direction.
    \newline    
    - \quad S represents AVATAR's location facing South direction.
    \newline
    \newline  
    Below I am giving you a solution code.
    \newline
    \newline    
    --- Solution ---
    \newline     
    \promptinput{\{solution\_code\}}
    \newline
    \newline
    Can you generate a task with 8x8 visual grid that would be solved by this code? The visual grid must contain AVATAR (with specific location and direction) along with GOAL, and can have WALL cells and FREE cells. Number your grid with row numbers (1 to 8) and column numbers (1 to 8). Also, you should tell me the position of AVATAR and GOAL in your generated task so we are sure about the numbering.
    \newline
    \newline
    You can verify the correctness of your generated task by executing the solution code on your task. A solution code for a task takes AVATAR to GOAL when executed. Note that AVATAR can only move on FREE cells and will crash if it tries to go to a WALL cell. If your generated task is not correct, you should try again to generate a correct task.
    \newline
    \newline
    --- Task ---
    \newline     
    \\
    \hline
\end{tabular}

    }
    \caption{Prompt for the task synthesis scenario in the \hocType{} domain. This prompt has several \promptinput{placeholders} to include details for the input solution code. Details are in Section~\ref{sec.tasksynthesis}.
    }
    \label{fig.task_synthesis.prompt_hoc}
\end{figure}

%
\paragraph{Output quality and performance metrics.} We assess the generated output along several quality attributes and use aggregated results over these quality attributes as performance metrics in our evaluation. All attributes for this scenario are binary, with a value of $1$ being better. \emph{LayoutCorrect} captures whether the general structure of the generated task is correct w.r.t. the underlying specification of tasks in the domain~\cite{neurtasksyn}. \emph{TaskSolvedByCode} captures whether the generated task is correctly solved by the input code. \emph{TaskSolvedByEdittedCode} is a relaxation of \emph{TaskSolvedByCode} criterion and captures whether the generated task can be solved after making a few edits to the input code. \emph{TaskSolvable} is a further relaxation of the solvability criteria and captures whether the task is solvable by any code. \emph{Overall} is $1$ when the following holds: (i) \emph{LayoutCorrect} attribute is $1$; (ii) at least one of the \emph{TaskSolvedByCode} or \emph{TaskSolvedByEdittedCode} attributes are $1$. Human evaluators manually annotate the quality of generated output for each of the $10$ instances as mentioned in Section~\ref{sec.problemsetup}.

\begin{figure*}[t!]
    \centering
    \scalebox{1.0}{
        \setlength\tabcolsep{2.5pt}
        \renewcommand{\arraystretch}{1.4}         
\begin{tabular}{l||c|cccc}
    \toprule
    \multicolumn{1}{c||}{\textbf{Method}} & \multicolumn{5}{c}{\textbf{Metrics}}\\
     & \multicolumn{1}{c|}{Overall} & \multicolumn{1}{c}{LayoutCorrect} & \multicolumn{1}{c}{TaskSolvedByCode} & \multicolumn{1}{c}{TaskSolvedByEdittedCode} & \multicolumn{1}{c}{TaskSolvable} \\  
    \midrule
    \TechChatGPT & $10.0$ & $70.0$ & $0.0$ & $10.0$ & $50.0$ \\
    \TechGPTFour & $20.0$ & $90.0$ & $0.0$ & $20.0$ & $80.0$ \\
    \bottomrule
\end{tabular}


    }
    \caption{Results for the task synthesis scenario. Details are in Section~\ref{sec.tasksynthesis}.}
    \label{fig.task_synthesis.results}
\end{figure*}

\begin{figure*}[t!]
\centering
    \begin{subfigure}[b]{0.49\textwidth}
        \centering
        \includegraphics[height=3.50cm]{./figs/2_problemsetup/files/fig_originalcode_hoc18.pdf}   
        \caption{Input: Solution code}
        \label{fig.task_synthesis.example.hoc.input} 
    \end{subfigure}
    \ \     
    \begin{subfigure}[b]{0.49\textwidth}
        \centering
        \includegraphics[height=3.44cm]{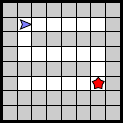}
        \caption{Output by \TechGPTFour{}}
        \label{fig.task_synthesis.example.hoc.output}
    \end{subfigure}
    %
    \caption{Illustrative example for the task synthesis scenario based on \emph{HoCMaze:18} task in Figure~\ref{fig.problemsetup.task.hoc}. This example highlights the struggles by \TechGPTFour{} in generating a task. \textbf{(a)} shows the solution code provided as input. \textbf{(b)} shows the task generated by \TechGPTFour{} as output. The generated task cannot be solved by the input code.
    }  
    \label{fig.task_synthesis.example.hoc}
    \vspace{-2mm}
\end{figure*}

\paragraph{Results.} Figure~\ref{fig.task_synthesis.results} provide results for various metrics aggregated across $10$ instances, reported in terms of \%. Next, we summarize some of the key findings. First, results in Figure~\ref{fig.task_synthesis.results} for the metric \emph{Overall} highlight that both \TechGPTFour{} ($20.0$) and \TechChatGPT{} ($10.0$) perform poorly for the task synthesis scenario. \TechGPTFour{} has slightly improved w.r.t. \TechChatGPT{} on the metrics of \emph{LayoutCorrect} and \emph{TaskSolvable}; however, it performs poorly as the generated task is not solvable by the input code. Second, these results highlight that \TechGPTFour{} struggles in generating visual programming tasks even for elementary-level codes with low complexity (see Figure~\ref{fig.problemsetup.dataset}). Figure~\ref{fig.solution_synthesis.example.hoc} provide an illustrative example highlighting the struggles by \TechGPTFour{} in generating a task for the solution code of \emph{HoCMaze:18} task.

%

\section{Concluding Discussions}\label{sec.conclusion}
We conducted a study to benchmark state-of-the-art generative AI and large language models in visual programming domains popularly used for K-8 programming education. Our results show that generative models like GPT-4 perform poorly in visual programming, in contrast to their advanced capabilities in text-based Python programming. In particular, our results highlight that these models struggle to combine spatial, logical, and programming skills crucial for visual programming. 

Next, we discuss some limitations of our current work and ideas to tackle them in the future. First, we considered only a small set of basic reference tasks from two visual programming domains; it would be interesting to conduct a study with a larger set that also comprises more complex reference tasks. Second, our performance assessment was based on a single generated output per instance; it would be useful to scale up the study where we evaluate multiple generated outputs per instance to account for the stochastic nature of these models.

Apart from the above extensions, there are many exciting directions for future work, including but not limited to: (a) curating novel benchmarks for visual programming that the research community can use to evaluate new versions of these models; (b) evaluating alternate generative models, in particular, open-source variants; (c) developing techniques to improve the performance of generative AI and large language models for visual programming, e.g., by leveraging symbolic methods, automated prompting, or fine-tuning.


\begin{ack}
Funded/Co-funded by the European Union (ERC, TOPS, 101039090). Views and opinions expressed are however those of the author(s) only and do not necessarily reflect those of the European Union or the European Research Council. Neither the European Union nor the granting authority can be held responsible for them.
\end{ack}

\bibliographystyle{unsrtnat}
\bibliography{main}

\clearpage
\begin{appendices}

\section*{Appendix}
This appendix provides prompts for the \karelType{} domain: (i) Figure~\ref{fig.execution_trace.prompt_karel} for the execution trace scenario; (ii) Figure~\ref{fig.solution_synthesis.prompt_karel} for the solution synthesis scenario; (iii) Figure~\ref{fig.task_synthesis.prompt_karel} for the task synthesis scenario.
\vspace{15mm}

\begin{figure}[h!]
    \centering
    \scalebox{0.90}{
        \setlength\tabcolsep{5pt}
        \renewcommand{\arraystretch}{1.2}
\begin{tabular}{||p{0.99\linewidth}||}
    \hline
    \multicolumn{1}{||c||}{\promptheader{Prompt: Execution Trace (\karelTypeBold{})}} \\ 
    I am learning to code using the visual programming domain of Karel programming.
    \newline
    \newline
    In this domain, the following types of coding blocks are available. 
    \newline
    - \quad Basic action blocks: move forward, turn left, turn right, pick marker, put marker.
    \newline
    - \quad Boolean conditions: path ahead, path to the left, path to the right, marker present, no path ahead, no marker present.
    \newline
    - \quad Loops: while(boolean)\{\}, repeat(int)\{\}.
    \newline
    - \quad Conditionals: If(boolean)\{\}, If(boolean)\{\}Else\{\}.
    \newline
    \newline
    In this domain, a task is represented as a pair of 10x10 visual pregrid and 10x10 visual postgrid. This pregrid and postgrid contain WALL cells, FREE cells, AVATAR (with specific location and direction), and markers. We represent a task's 10x10 visual pregrid and postgrid with the following symbols.
    \newline  
    - \quad \# represents a WALL cell.
    \newline
    - \quad + represents a FREE cell.
    \newline    
    - \quad m represents a cell with marker.
    \newline    
    - \quad E represents AVATAR's location on a cell without marker, facing East direction.
    \newline    
    - \quad W represents AVATAR's location on a cell without marker, facing West direction.
    \newline    
    - \quad N represents AVATAR's location on a cell without marker, facing North direction.
    \newline    
    - \quad S represents AVATAR's location on a cell without marker, facing South direction.
    \newline    
    - \quad Em represents AVATAR's location on a cell with marker, facing East direction.
    \newline    
    - \quad Wm represents AVATAR's location on a cell with marker, facing West direction.
    \newline    
    - \quad Nm represents AVATAR's location on a cell with marker, facing North direction.
    \newline    
    - \quad Sm represents AVATAR's location on a cell with marker, facing South direction.
    \newline
    \newline  
    Below I am giving you a task and its solution code. A solution code for a task transforms the pregrid into the postgrid when executed.
    \newline
    \newline    
    --- Task: Pregrid ---
    \newline    
    \promptinput{\{pregrid\_ascii\_representation\}}    
    \newline
    \newline
    --- Task: Postgrid ---
    \newline
    \promptinput{\{postgrid\_ascii\_representation\}}    
    \newline
    \newline
    --- Solution ---
    \newline    
    \promptinput{\{solution\_code\}}
    \newline
    \newline
    Can you produce an execution trace of this code on the task and tell me the sequence of AVATAR's positions, i.e., location and direction? Recall that a solution code for a task transforms the pregrid into the postgrid when executed.  In this task, AVATAR's position in the pregrid is ((row=\promptinput{\{avatar\_pre\_row\}}, col=\promptinput{\{avatar\_pre\_col\}}), \promptinput{\{avatar\_pre\_dir\}}), and AVATAR's position in the postgrid is ((row=\promptinput{\{avatar\_post\_row\}}, col=\promptinput{\{avatar\_post\_col\}}), \promptinput{\{avatar\_post\_dir\}}). Note that AVATAR can only move on FREE cells and will crash if it tries to go to a WALL cell. 
    \\
    \hline
\end{tabular}

    }
    \caption{Prompt for the execution trace scenario in the \karelType{} domain. This prompt has several \promptinput{placeholders} to include details for the input task and solution code. Details are in Section~\ref{sec.executiontrace}.
    }
    \label{fig.execution_trace.prompt_karel}
\end{figure}


\clearpage

\begin{figure}[t!]
    \centering
    \scalebox{0.90}{
        \setlength\tabcolsep{5pt}
        \renewcommand{\arraystretch}{1.2}
\begin{tabular}{||p{0.99\linewidth}||}
    \hline
    \multicolumn{1}{||c||}{\promptheader{Prompt: Solution Synthesis (\karelTypeBold{})}} \\ 
    I am learning to code using the visual programming domain of Karel programming.
    \newline
    \newline
    In this domain, the following types of coding blocks are available. 
    \newline
    - \quad Basic action blocks: move forward, turn left, turn right, pick marker, put marker.
    \newline
    - \quad Boolean conditions: path ahead, path to the left, path to the right, marker present, no path ahead, no marker present.
    \newline
    - \quad Loops: while(boolean)\{\}, repeat(int)\{\}.
    \newline
    - \quad Conditionals: If(boolean)\{\}, If(boolean)\{\}Else\{\}.
    \newline
    \newline
    In this domain, a task is represented as a pair of 10x10 visual pregrid and 10x10 visual postgrid. This pregrid and postgrid contain WALL cells, FREE cells, AVATAR (with specific location and direction), and markers. We represent a task's 10x10 visual pregrid and postgrid with the following symbols.
    \newline  
    - \quad \# represents a WALL cell.
    \newline
    - \quad + represents a FREE cell.
    \newline    
    - \quad m represents a cell with marker.
    \newline    
    - \quad E represents AVATAR's location on a cell without marker, facing East direction.
    \newline    
    - \quad W represents AVATAR's location on a cell without marker, facing West direction.
    \newline    
    - \quad N represents AVATAR's location on a cell without marker, facing North direction.
    \newline    
    - \quad S represents AVATAR's location on a cell without marker, facing South direction.
    \newline    
    - \quad Em represents AVATAR's location on a cell with marker, facing East direction.
    \newline    
    - \quad Wm represents AVATAR's location on a cell with marker, facing West direction.
    \newline    
    - \quad Nm represents AVATAR's location on a cell with marker, facing North direction.
    \newline    
    - \quad Sm represents AVATAR's location on a cell with marker, facing South direction.
    \newline
    \newline  
    Below I am giving you a task as a pair of 10x10 visual pregrid and 10x10 visual postgrid.
    \newline
    \newline    
    --- Task: Pregrid ---
    \newline    
    \promptinput{\{pregrid\_ascii\_representation\}}    
    \newline
    \newline
    --- Task: Postgrid ---
    \newline
    \promptinput{\{postgrid\_ascii\_representation\}}    
    \newline
    \newline
    Can you generate a solution code for this task that uses the minimum number of blocks? A solution code for a task transforms the pregrid into the postgrid when executed. In this task, AVATAR's position in the pregrid is ((row=\promptinput{\{avatar\_pre\_row\}}, col=\promptinput{\{avatar\_pre\_col\}}), \promptinput{\{avatar\_pre\_dir\}}), and AVATAR's position in the postgrid is ((row=\promptinput{\{avatar\_post\_row\}}, col=\promptinput{\{avatar\_post\_col\}}), \promptinput{\{avatar\_post\_dir\}}). Note that AVATAR can only move on FREE cells and will crash if it tries to go to a WALL cell.
    \newline
    \newline    
    --- Solution ---
    \newline    
    \\
    \hline
\end{tabular}

    }
    \caption{Prompt for the solution synthesis scenario in the \karelType{} domain. This prompt has several \promptinput{placeholders} to include details for the input task. Details are in Section~\ref{sec.solutionsynthesis}.
    }
    \label{fig.solution_synthesis.prompt_karel}
\end{figure}


\clearpage

\begin{figure}[t!]
    \centering
    \scalebox{0.90}{
        \setlength\tabcolsep{5pt}
        \renewcommand{\arraystretch}{1.2}
\begin{tabular}{||p{0.99\linewidth}||}
    \hline
    \multicolumn{1}{||c||}{\promptheader{Prompt: Task Synthesis (\karelTypeBold{})}} \\ 
    I am learning to code using the visual programming domain of Karel programming.
    \newline
    \newline
    In this domain, the following types of coding blocks are available. 
    \newline
    - \quad Basic action blocks: move forward, turn left, turn right, pick marker, put marker.
    \newline
    - \quad Boolean conditions: path ahead, path to the left, path to the right, marker present, no path ahead, no marker present.
    \newline
    - \quad Loops: while(boolean)\{\}, repeat(int)\{\}.
    \newline
    - \quad Conditionals: If(boolean)\{\}, If(boolean)\{\}Else\{\}.
    \newline
    \newline
    In this domain, a task is represented as a pair of 10x10 visual pregrid and 10x10 visual postgrid. This pregrid and postgrid contain WALL cells, FREE cells, AVATAR (with specific location and direction), and markers. We represent a task's 10x10 visual pregrid and postgrid with the following symbols.
    \newline  
    - \quad \# represents a WALL cell.
    \newline
    - \quad + represents a FREE cell.
    \newline    
    - \quad m represents a cell with marker.
    \newline    
    - \quad E represents AVATAR's location on a cell without marker, facing East direction.
    \newline    
    - \quad W represents AVATAR's location on a cell without marker, facing West direction.
    \newline    
    - \quad N represents AVATAR's location on a cell without marker, facing North direction.
    \newline    
    - \quad S represents AVATAR's location on a cell without marker, facing South direction.
    \newline    
    - \quad Em represents AVATAR's location on a cell with marker, facing East direction.
    \newline    
    - \quad Wm represents AVATAR's location on a cell with marker, facing West direction.
    \newline    
    - \quad Nm represents AVATAR's location on a cell with marker, facing North direction.
    \newline    
    - \quad Sm represents AVATAR's location on a cell with marker, facing South direction.
    \newline
    \newline  
    Below I am giving you a solution code. 
    \newline
    \newline 
    --- Solution ---
    \newline    
    \promptinput{\{solution\_code\}}
    \newline
    \newline
    Can you generate a task with a pair of 10x10 visual pregrid and 10x10 visual postgrid that would be solved by this code? Both the visual pregrid and visual postgrid must contain AVATAR (with specific location and direction), and can have WALL cells, FREE cells, and markers. Number your grids with row numbers (1 to 10) and column numbers (1 to 10). Also, you should tell me the position of AVATAR in your generated pregrid and postgrid so we are sure about the numbering.
    \newline
    \newline
    You can verify the correctness of your generated task by executing the solution code on your task. A solution code for a task transforms the pregrid into the postgrid when executed. Note that AVATAR can only move on FREE cells and will crash if it tries to go to a WALL cell. If your generated task is not correct, you should try again to generate a correct task.
    \newline
    \newline
    --- Task ---
    \newline
    \\
    \hline
\end{tabular}

    }
    \caption{Prompt for the task synthesis scenario in the \karelType{} domain. This prompt has several \promptinput{placeholders} to include details for the input solution code. Details are in Section~\ref{sec.tasksynthesis}.
    }
    \label{fig.task_synthesis.prompt_karel}
\end{figure}


\clearpage


\end{appendices}

\end{document}